\documentclass[letterpaper, 10 pt, conference]{ieeeconf}  

\IEEEoverridecommandlockouts                              

\usepackage{times}
\usepackage{graphicx}
\usepackage{booktabs}
\usepackage{epsfig} 
\usepackage{amsmath} 
\usepackage{amssymb}  
\usepackage{breqn}
\usepackage{cite}
\usepackage{bm}
\usepackage{multicol, blindtext}
\usepackage{makecell}
\usepackage{float}
\usepackage{siunitx}
\usepackage{xcolor}
\usepackage{url}
\usepackage{hyperref}
\usepackage{lscape} 
\usepackage{multirow}
\usepackage{graphics}
\usepackage{pdfpages}

\DeclareGraphicsExtensions{.png,.pdf}

\title{\LARGE \bf

Learning to Grasp Clothing Structural Regions\\ for Garment Manipulation Tasks
}

\author{~Wei~Chen, Dongmyoung~Lee, Digby Chappell,
        and~Nicolas~Rojas
\thanks{Wei Chen, Dongmyoung Lee, Digby Chappell,
        and Nicolas Rojas are with the REDS Lab, Dyson School of Design Engineering, Imperial College London, 25 Exhibition Road, London, SW7 2DB, UK
{\tt\small (w.chen21, d.lee20, d.chappell19, n.rojas)@imperial.ac.uk}}
\thanks{This work was supported in part by the China Scholarship Council, and in part by the Dyson School of Design Engineering, Imperial College London. Digby Chappell was supported by the UKRI CDT in AI for Healthcare \protect\url{http://ai4health.io} (Grant No. EP/S023283/1).}
}
\begin{document}
\maketitle

\thispagestyle{empty}
\pagestyle{empty}

\begin{abstract}
When performing cloth-related tasks, such as garment hanging, it is often important to identify and grasp certain structural regions---a shirt's collar as opposed to its sleeve, for instance. However, due to cloth deformability, these manipulation activities, which are essential in domestic, health care, and industrial contexts, remain challenging for robots. In this paper, we focus on how to segment and grasp structural regions of clothes to enable manipulation tasks, using hanging tasks as case study. To this end, a neural network-based perception system is proposed to segment a shirt's collar from areas that represent the rest of the scene in a depth image.
With a 10-minute video of a human manipulating shirts to train it, our perception system is capable of generalizing to other shirts regardless of texture as well as to other types of collared garments. A novel grasping strategy is then proposed based on the segmentation to determine grasping pose. Experiments demonstrate that our proposed grasping strategy achieves 92\%, 80\%, and 50\% grasping success rates with one folded garment, one crumpled garment and three crumpled garments, respectively. 
Our grasping strategy performs considerably better than tested baselines that do not take into account the structural nature of the garments.
With the proposed region segmentation and grasping strategy, challenging garment hanging tasks are successfully implemented using an open-loop control policy. Supplementary material is available at \href{https://sites.google.com/view/garment-hanging}{https://sites.google.com/view/garment-hanging}

\end{abstract}

\section{Introduction}



Manipulating items of clothing is a task that is critical to a broad spectrum of tasks, ranging from domestic chores, assistive dressing, and industrial automation.
%
Humans are extremely gifted at tasks such as these, due to the complex perception and sensorimotor systems present in our bodies enabling precise, robust, and delicate manipulation of many objects~\cite{sobinov2021neural}.
Although significant progress in the robotic manipulation of rigid body objects has been realized, due to the deformable nature of many garments, robotic manipulation of clothing still remains an open challenge.
Robotic manipulators lag far behind the human hand in terms of controllable degrees of freedom and sensory capabilities, so an question is raised as to what level of clothing manipulation is possible with existing technology, and how \textit{a priori} knowledge of the structure of garments can be utilized to supplement these missing capabilities.
%

Clothing design elements are critical structures that affect the general configuration of clothing (e.g. collar of a shirt, waistband of pants) \cite{hong2008eliciting}.
Perception of such clothing structural regions is of particular importance for robot manipulation tasks, none more so than when attempting to manipulate deformable and continuum objects \cite{jimenez2020perception}. Without the ability to detect clothing structural regions, downstream manipulation becomes inherently limited. For example, folding or hanging clothes becomes increasingly difficult without knowledge of suitable grasping locations on the garment.
Many prior works on the perception of generic cloth for manipulation have primarily focused on detecting edges, wrinkles, or corners of the material in order to achieve a successful grasp \cite{qian2020cloth, sun2015accurate, maitin2010cloth}. Both classical computer vision (CV)-based and learning-based methods were applied in finding these edged or wrinkled features.
While grasp success is increased via these methods, such works are only able to utilize the grasps in the form of relatively simple manipulation tasks such as cloth folding and unfolding. For more complex tasks such as garment-hanging, it would be challenging to apply these methods successfully without further knowledge of the garment.


\begin{figure}[t!]
    \centering
    \vspace{4mm}
    \includegraphics[width=0.98\columnwidth]{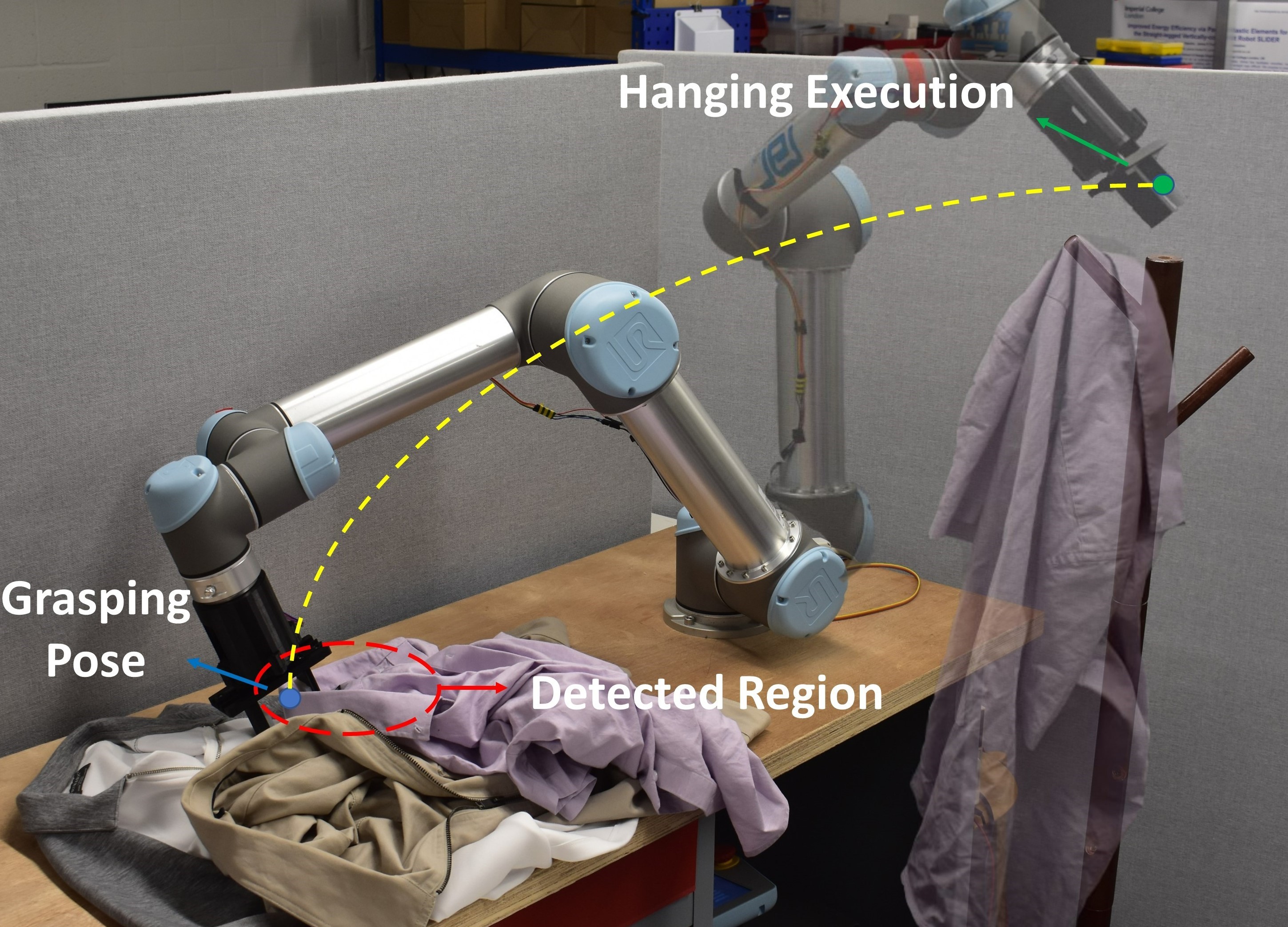}
    \caption{In the proposed method, a robot learns to automatically detect a clothing structural region (collar in our case study) in severely crumpled garments and execute its grasping. This is then used for cloth manipulation (hanging the garment on a cloth tree in our case study).}
    \label{fig:intro}
    \vspace{-10pt}
\end{figure}

This work presents a novel system for the robust robotic manipulation of garments by leveraging structural elements of these garments during perception and grasping as shown in Fig. \ref{fig:intro}. This system is tested extensively on the manipulation of collared-garments such as shirts and jackets for a novel garment-hanging task. 

In order to successfully grasp structural areas of the garments a data acquisition technique and deep learning-based segmentation model are required, in which a 10-minute video of unstructured human manipulation of multiple color-tagged garments is used to quickly extract ground truth segmentation masks for training. A novel grasping strategy is then proposed that outputs a suitable grasp pose based on the extracted skeleton and local geometric structure of the segmented region of clothing. Finally, the suitability of the grasping strategy for downstream manipulation tasks is tested on a garment-hanging task.
Our findings indicate that, by considering only the structural regions of the garment that are vital for task success, robust clothing manipulation can be achieved with standard perception and robotic hardware.

\section{Related Work}
\subsection{Cloth Perception}
In order for cloth manipulation to be realized, perception of key elements of the cloth is vital. Several approaches have contributed to the area of cloth perception. Due to the deformable characteristics of cloth, wrinkles~\cite{sun2017single, yuan2018active}, folds~\cite{moriya2018method}, and corners~\cite{seita_bedmake_2019, jangir2020dynamic, weng2022fabricflownet,matas2018sim} are generally considered to be attractive grasping targets for manipulation tasks such as cloth folding or flattening.
Accordingly, early works of these approaches based on traditional CV algorithms aimed to detect these cloth features. Typically, after some image processing, corner and edge detection methods have been used to classify cloths~\cite{willimon2011classification} or locate interaction points for cloth flattening tasks~\cite{maitin2010cloth, triantafyllou2011vision}. In addition, \cite{sun2015accurate} used shape index to analyse the topology of the garment surface to detect wrinkles. More recent progress in cloth perception has focused on learning-based methods to achieve the same goal. Notable works include \cite{seita_bedmake_2019}, which applied YOLO with pre-trained weights to identify corner points for robotic bed making, and \cite{weng2022fabricflownet}, which combined optical flow with a neural network to learn grasping points for cloth folding.

Although detecting the corners and edges of a cloth is useful for basic tasks such as flattening, the lack of semantic meaning makes it difficult to be applied to tasks where structural elements of the cloth are important, as is the case in many tasks involving items of clothing. This increases the complexity of the problem twofold, as the perception of a structured garment is more difficult, and also because the task and context in which the garment must be perceived is likely to become more complicated as well. Some works have focused on understanding the structure of the clothes in order to facilitate a better grasp, attempting to reconstruct the garment in 3D canonical space~\cite{chi2021garmentnets}. However, in the training of 3D reconstruction methods, challenges arise regarding computational complexity and data acquisition, since the training of such deep learning models requires high computational power and high-quality manually labelled point cloud datasets.

\subsection{Cloth Grasping for Manipulation}
During cloth grasping, the manipulation task to be performed provides context that can either simplify or add complexity to the grasping problem~\cite{jimenez2020perception}. For example, a robot aiming to load laundry into a washing machine need not perceive structural features of the clothes it is manipulating~\cite{shehawy2021estimating}, whereas a robot that is assisting dressing must consistently and accurately grasp specific points on the item of clothing that is to be worn~\cite{saxena2019garment, zhang2020learning}. For tasks that do not consider the overall structure of the target clothes, some generic features are preferred to be selected as the grasping point. For instance, many simple cloth folding and unfolding setups have perception systems that are devoted to locating previously mentioned generic features, such as edge or corner points~\cite{maitin2010cloth, triantafyllou2011vision, weng2022fabricflownet}.

One of the largest challenges when grasping more complex items of clothing is to estimate the 6D pose that the gripper must reach in order to achieve a successful grasp. This is a complex perception task, requiring detailed knowledge of the structure of the clothing, the current state of the clothing, and the manipulation task to be performed. Rather than simplifying the garment or the manipulation task itself, some works have conveniently simplified the initial state of the clothing in order to facilitate a simpler grasping strategy. This is typically achieved by semi-constraining the grasping pose to an approximately planar setup~\cite{ha2022flingbot, seita_bedmake_2019, matas2018sim}, or by having the clothes pre-hung on a peg to allow gravity to naturally restrict the orientation of possible grasping points~\cite{saxena2019garment, zhang2020learning, zhang2022learning}. Furthermore, the vast majority of works that focus on complex manipulation tasks only focus on a single garment at a time.

There is a notable lack of research investigating clothing manipulation from more complex initial setups. In this work, we investigate the problem of hanging shirts on a peg-based hanger from an unrestricted or crumpled initial state. This is a relatively precise manipulation task requiring knowledge of the structure of the garment. Furthermore, we explore the effect of the initial setup on the ability of our proposed system to perform the task by testing initial configurations with multiple folded and crumpled shirts.

\begin{figure*}[t!]
    \centering
    \vspace{4mm}
    \includegraphics[width=0.98\textwidth]{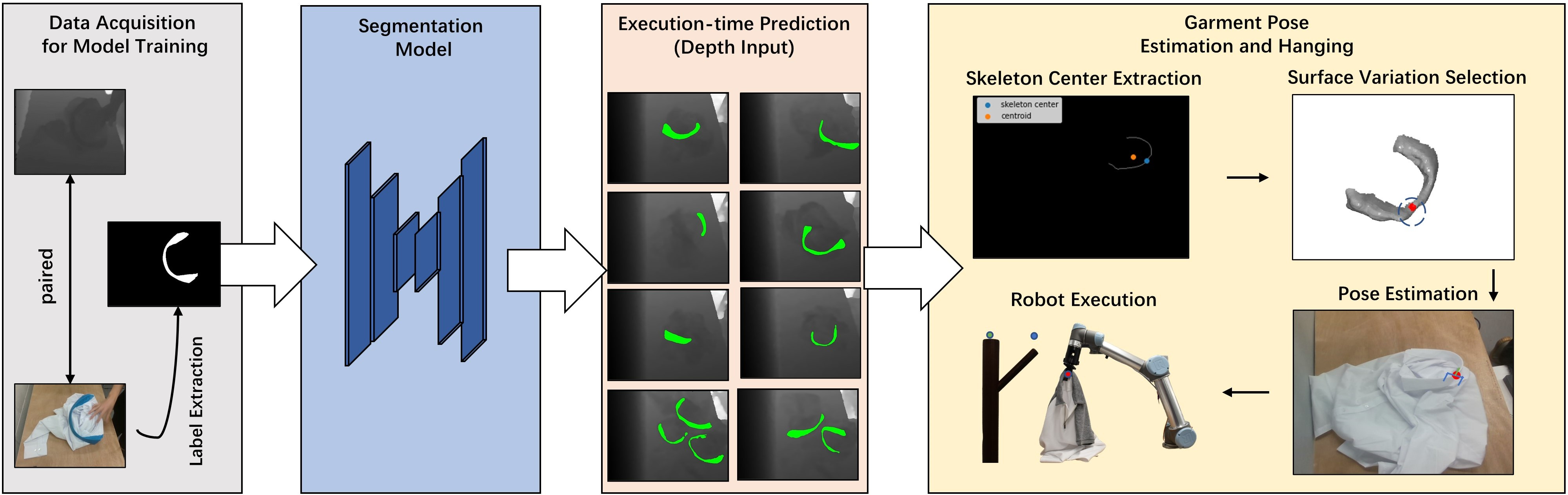}
    \caption{Pipeline of Our Method: We collect depth image and its corresponding RGB image of the diverse configuration of garments.  The collected depth images are labelled by extracting the blue pixels from RGB images and used as the training set.  During the running time, We use the depth image as input to predict the collar by the trained segmentation neural network;  The optimal grasping point is obtained by fitting a skeleton to the predicted collar and calculating the surface variation.  A local PCA is finally conducted near this point to find the grasping orientation.  A set of action primitives, including grasping and hanging, are designed for performing real-world robot execution.}
    \label{fig:data_acquisition}
        \vspace{-10pt}
\end{figure*}
\section{Methods}
Successful manipulation always requires the garment to be grasped in a specific region. In this paper, we propose a novel pipeline that detects the collar region of crumpled garments to perform garment hanging. To this end, a deep neural network is applied for the perception system. One of the most significant challenges of training deep neural networks is data acquisition. We propose a
data collection process which uses video clips that record human manipulating cloth for large dataset generation. This dataset can be directly fed into the neural network for training with appropriate pre-processing techniques.
With the trained neural network, the collar region of garments can be then segmented. Within the collar region, we estimate the most graspable point to provide robust robotic cloth grasping. Specifically, we extract the collar region's center using a skeleton-based method. Then, the point with the highest surface variation within the center region is finally chosen as the grasping point, as this represents the sharp fold of the collar. Principal component analysis (PCA) is applied for the pose estimation of the grasping point. The insertion grasping policy and hanging execution are finally conducted for the robotic cloth grasping and hanging. The overall system is shown in Fig. \ref{fig:data_acquisition}.

\subsection{Cloth Region Segmentation}
This section presents a deep neural network-based perception system that detects the collar part of severely crumpled garments placed on a tabletop. The neural network is trained on video clips recorded from a human freely manipulating garments. The whole perception pipeline is illustrated in Fig. \ref{fig:data_acquisition}, including data acquisition and model training.

\subsubsection{Data Acquisition}
While training a deep neural network requires a huge number of datasets, such datasets are not readily available. The traditional data acquisition method requires much human effort for manual annotation \cite{ramisa2012using}. While using a physics engine to generate synthetic pictures to enrich the hybrid datasets will also cause the sim-to-real gap \cite{zhang2020learning}.
We instead apply a mechanism similar to \cite{qian2020cloth} to obtain data directly from the real world.

\begin{figure}[t!]
    \centering
    \vspace{4mm}
    \includegraphics[width=0.98\columnwidth]{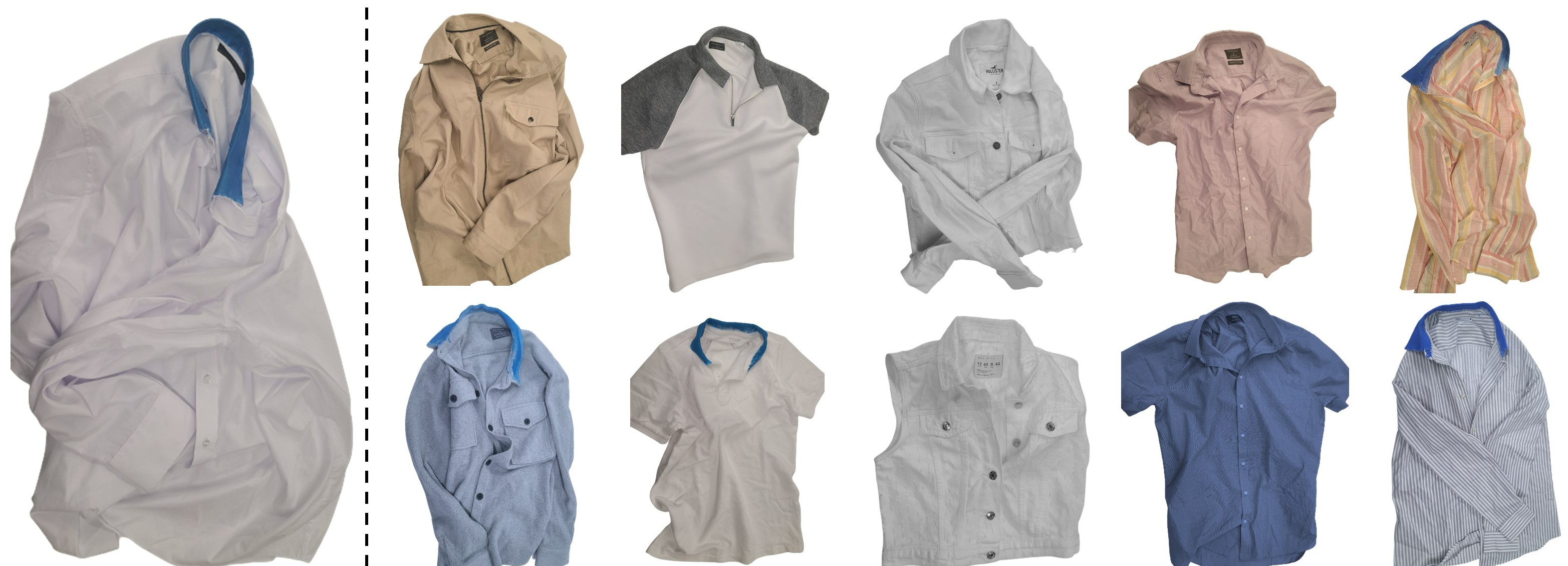}
    \caption{\textbf{Left}: template shirts used for perception system training in Section III-A. \textbf{Right}: Various garments were used in the testing section of our study. The collar is pained with blue for groundtruth extraction.}
    \label{fig:garments}
    \vspace{-10pt}
\end{figure}

As shown in Fig. \ref{fig:garments}, our data acquisition process uses white template shirts (TSs) with long sleeves. We paint the collar of TSs to be blue. A human operator randomly manipulates these TSs with an RGB-D camera fixed on the top, simultaneously capturing RGB and depth images. The camera frame rate is set to 15Hz to ensure the TSs among each frame have appropriate deformation. We also include multiple pieces (maximum three) of shirts to increase the clutter level rather, than just one shirt during the data collection procedure. This dataset frame order is then shuffled before being fed to the neural network to break the temporal correspondence and make the dataset uniformly distributed. After the shuffling, the ground truth can be obtained by extracting the blue colour from the RGB images in HSV space. These RGB images are only used for labelling and are not in the training procedure or for further segmentation.



\subsubsection{Semantic Segmentation with Neural Networks}
We apply a deep neural network based on U-net \cite{ronneberger2015u}. Resnet-50 block \cite{he2016deep} with pre-trained weights on imagenet\cite{deng2009imagenet} is applied as the encoder to extract image features. Due to the fact that the resnet-50 block uses weights pre-trained on RGB pictures, we stack depth images three times to match the three channels of RGB. We formulate the semantic segmentation as a pixel-level binary classification task that only predicts the presence of collar pixels to exclude other features in the collected depth image (e.g. tabletop, human hand).
Dice loss is applied here to reduce the effect of imbalanced class distribution. Data augmentation procedures such as flipping, rotation and grid distortion are also implemented to avoid model over-fitting.

\subsection{Grasping Pose Estimation}
Choosing a centralized area with a wrinkled surface will give a higher probability of successful grasp \cite{jimenez2020perception}. Rather than calculating graspable region using a complex hierarchical vision architecture, such as one defined from shape index in \cite{sun2015accurate}, here, an effective point cloud-based grasping pose estimation algorithm is proposed. This proposed grasping strategy involves three stages: center region extraction, grasping point selection, and orientation estimation. 

\subsubsection{Center Region Extraction}
\begin{figure}[t!]
    \centering
    \vspace{4mm}
    \includegraphics[width=0.98\columnwidth]{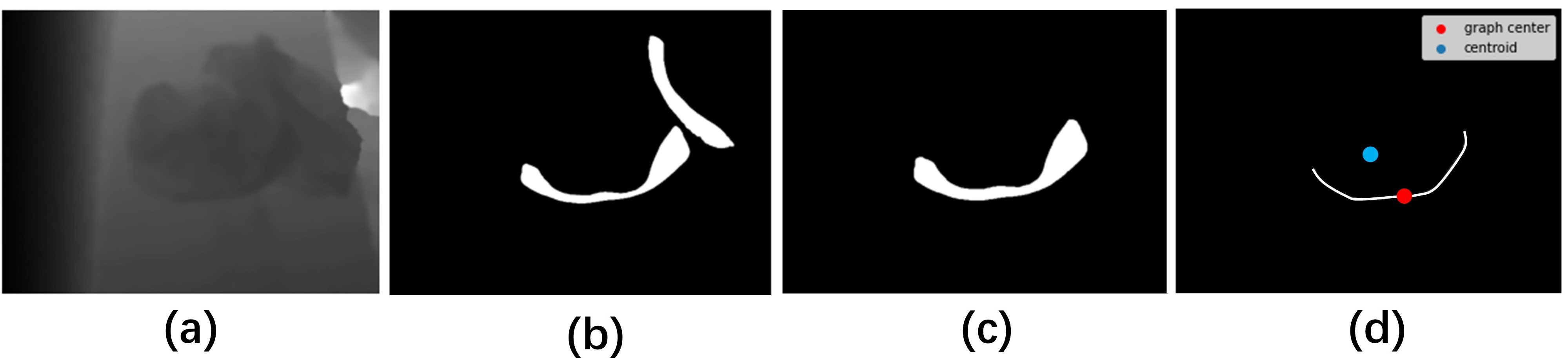}
    \caption{Center Region Extraction: (a) Input depth image; (b) Prediction result; (c) Clustering; (d) Skeletonize and calculate the center of skeleton. We compare the centriod (blue dot) and skeleton center (red dot).}
    \label{fig:extraction}
    \vspace{-10pt}
\end{figure}

Grasping at the extreme edges of the collar may cause empty grasps or slippage due to a smaller contact area between the gripper and the collar.
To avoid this, the center region of cloth is targeted, ensuring a stable grasping action due to the larger contact area.
However, the centroid or the center of the convex hull may not provide a good candidate for grasping concave shape garments, especially for the collar of garments which are typically concave. Inspired by \cite{jimenez2020perception}, we adopt a skeleton-based method to estimate the middle point of the skeleton, therefore extracting the center region of the collar. Three steps are involved in our center region extraction algorithm, shown in Fig. \ref{fig:extraction}:
\begin{itemize}
\item
\textbf{Clustering}:
After obtaining the prediction result, we first apply agglomerative clustering to the prediction result based on the pixel locations to identify individual collars from a scene potentially containing multiple garments. The largest cluster is chosen for the following grasping pipeline. After the largest cluster is obtained, an image dilation is applied to remove noise from the clustered prediction result.
\end{itemize}

\begin{itemize}
\item
\textbf{Skeletonize}:
The skeletonize algorithm is then applied to this largest cluster. This skeletonize algorithm is originally from \cite{zhang1984fast}, and applied to the prediction result to obtain a boundary edge map which is a single pixel wide, in other words, a skeleton.
\end{itemize}

\begin{itemize}
\item
\textbf{Center extraction}:
Closeness centrality is defined by calculating the inverse of the total sum of shortest distances from a node to every other node \cite{freeman1978centrality}. We find the maximum closeness centrality here to represent the midpoint of the whole skeleton. This pixel is selected as the center point of the collar.
\end{itemize}

\subsubsection{Grasping Position Selection}

Using the extracted collar pixels, we de-project these collar pixels from the depth image to obtain the collar point cloud according to the known camera intrinsic parameters. Pre-processing procedures such as voxel down-sampling and radius outlier removal are implemented here to reduce computational complexity\cite{Zhou2018}.
Since cloth is highly deformable, a random point selection near the skeleton center is insufficient to achieve a robust clothing grasp. It has been shown that wrinkled points are good candidates for the gripping action; successful grasping can be guaranteed by positioning the grippers at either side of a crease \cite{jimenez2020perception}. To ensure a successful and stable grasp, wrinkles with more significant curvature changes are favoured. We use surface variation to approximate these curvature changes. Having the predicted collar point cloud and estimated skeleton center, surface variation estimation near the skeleton center of the local point cloud is applied \cite{pauly2002efficient}.

For each point in the region of $N$ nearest points (here chosen to be $N=50$) to the skeleton center $P$, we compute local surface variation. Surface variation for point $\bm{p}^{(i)}$ is computed by selecting $n$ neighbouring points, $[\bm{p}^{(i)}_1, \ldots, \bm{p}^{(i)}_n]^T$ and first calculating the centroid position $\Bar{\bm{\mu}}^{(i)}$:
\begin{equation}
    \Bar{\bm{\mu}}^{(i)} = \frac{1}{n}\sum_{j=1}^{n} \bm{p}^{(i)}_j,
\end{equation}
where $n=50$ as in \cite{pauly2002efficient}. Then, the covariance matrix $\bm{C}^{(i)}$ is obtained:
\begin{equation}
    \bm{C}^{(i)} = \frac{1}{n}\sum_{j=1}^{n} (\bm{p}^{(i)}_j - \Bar{\bm{\mu}}^{(i)})(\bm{p}^{(i)}_j - \Bar{\bm{\mu}}^{(i)})^T.
\end{equation}
The eigenvectors and eigenvalues of $\bm{C}^{(i)}$ are computed:
\begin{equation}
    \bm{C}^{(i)}\overrightarrow{\bm{v}}^{(i)}_k = \lambda_k\overrightarrow{\bm{v}}^{(i)}_k, \quad k \in\{0,1,2\},
\end{equation}
where the eigenvalue $\lambda_{k}$ quantifies the variance in the direction of eigenvector $\overrightarrow{\bm{v}}^{(i)}_{k}$ at point $\bm{p}^{(i)}$. Assuming the three eigenvalues are ordered, $\lambda_{0}<\lambda_{1}<\lambda_{2}$, the eigenvector $\overrightarrow{\bm{v}}^{(i)}_{0}$ then represents the axis normal to the direction of maximum surface variation, and $\overrightarrow{\bm{v}}^{(i)}_{2}$ to the axis of maximum surface variation \cite{rusu2008towards}.
The three eigenvalues of the covariance matrix $\bm{C}$ are used to approximate the local surface variation around point $\bm{p}^{(i)}$. The surface variation at point $\bm{p}^{(i)}$ can be obtained, as in \cite{pauly2002efficient}, by computing:
\begin{equation}
    \sigma_{\bm{p}^{(i)}}=\frac{\lambda_{0}}{\lambda_{0}+\lambda_{1}+\lambda_{2}}.
\end{equation}
We compute the surface variation at each of the $N$ neighboring points around skeleton center $P$, and select the point of largest surface variation, $\bm{p}$, within the neighborhood as the grasping point, and its own $n$ nearest neighbors as the grasping region.

 \begin{figure}[t!]
    \centering
     \vspace{4mm}
    \includegraphics[width=0.98\columnwidth]{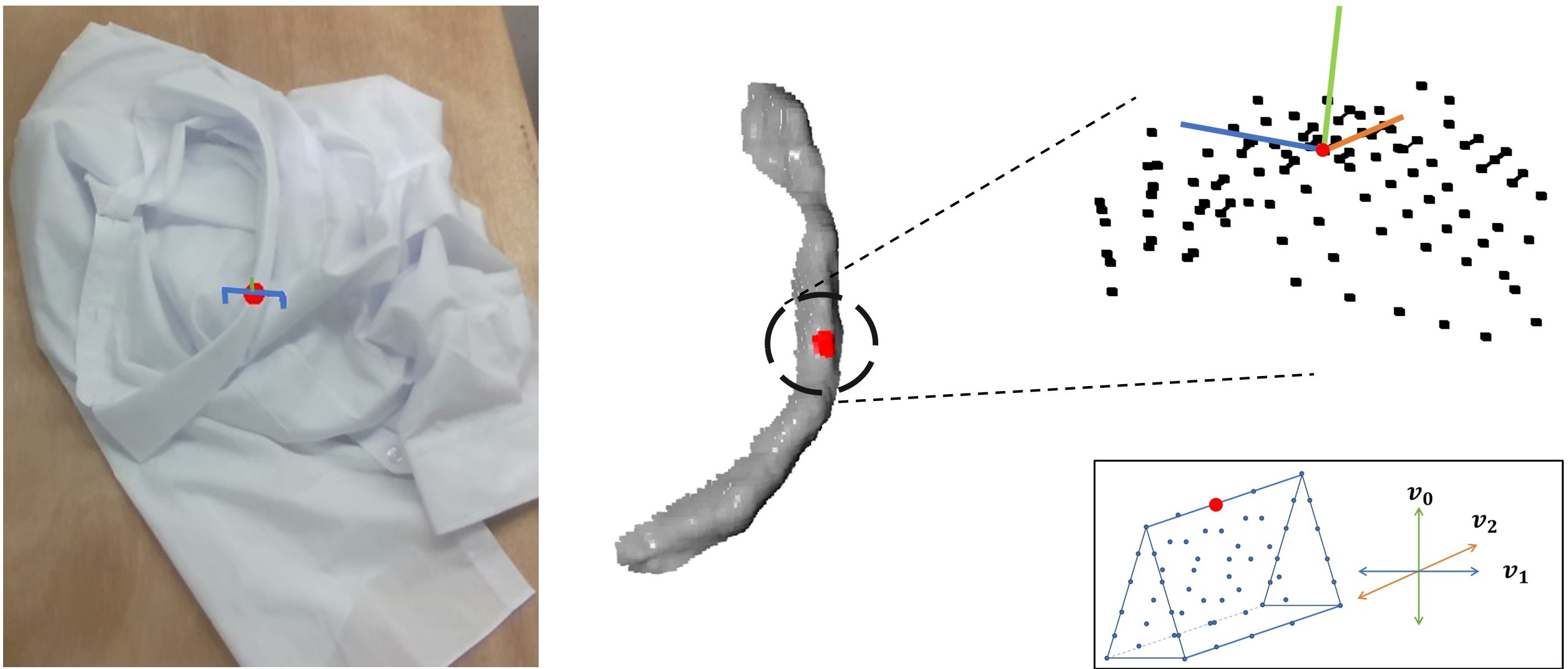}
    \caption{Grasping Pose Estimation: garment RGB image and extracted collar region pointcloud. The point (red point) with highest surface variation in the center region (dashed circle) will be selected as the grasping point. Three eigenvectors ($\overrightarrow{\bm{v}}_{0}$, $\overrightarrow{\bm{v}}_{1}$, $\overrightarrow{\bm{v}}_{2}$) from this point and its nearest neighbour are used for the pose estimation.}
    \label{fig:pose}
    \vspace{-10pt}
\end{figure}

\begin{table*}[t!]
   \centering
    \vspace{4mm}
   \caption{Perception Evaluation Results of Seen and Unseen Garments}\label{tab:perception}
        \vspace{-2mm}
      \setlength{\tabcolsep}{0.9mm}{
   \begin{tabular}{ c|c c c c c c c c c c c}
       \hline
        & TS (seen) & S1 (unseen) & S2 (unseen) & L1 (unseen) & L2 (unseen)& P1(unseen) &P2(unseen)& C1(unseen) &C2(unseen)& D1 (unseen)& D2 (unseen)\\ [0.20ex]
        \hline
        {Recall}      &0.873 &0.799 &0.665 &0.778 &0.751 &0.304 &0.603 &0.623 &0.585 &0.681 &0.772\\
        {Precision}   &0.853 &0.825 &0.863 &0.792 &0.791 &0.636 &0.808 &0.925 &0.838 &0.891 &0.865\\
        {IoU}         &0.760 &0.684 &0.602 &0.645 &0.626 &0.255 &0.527 &0.593 &0.525 &0.628 &0.689\\
        \hline

   \end{tabular}}
   \par
    \begin{flushleft}
Evaluation results of our perception system. Our perception system is trained by dataset collected from template shirts (TS) and evaluated by testing dataset from unseen garments: short sleeve shirts (S1, S2), long sleeve shirts (L1, L2), polo shirts (P1, P2), coats (C1, C2) and denim jackets (D1, D2).
    \end{flushleft}
    \vspace{-10pt}
\end{table*}

\subsubsection{Grasping Orientation Estimation}

The grasping pose is defined by aligning the gripper pose to the grasping point for the actual robotic cloth grasping instead of estimating the whole pose of the cloth \cite{zhang2020learning}.
We use the eigenvector $\overrightarrow{\bm{v}}_{0}$ (correlates with the smallest eigenvalue), which is estimated from the mean covariance matrix as the surface normal vector $\bm{n}$ of the selected grasping region. This surface normal vector $\bm{n}$ is then defined as the $Z$-axis of the orientation. As shown in illustration in Fig. \ref{fig:pose}, the major component of eigenvector $\overrightarrow{\bm{v}}_{2}$ is aligned with the average longitudinal direction of the selected grasping region in which we define the $Y$-axis here. Since these three eigenvectors  $\overrightarrow{\bm{v}}_{0}$, $\overrightarrow{\bm{v}}_{1}$, $\overrightarrow{\bm{v}}_{2}$ are orthogonal to each other, the $X$-axis will be aligned with eigenvector $\overrightarrow{\bm{v}}_{1}$ at the lateral direction.

 \subsection{Insertion Grasping Execution}
\begin{figure}[t!]
    \centering
    \includegraphics[width=0.98\columnwidth]{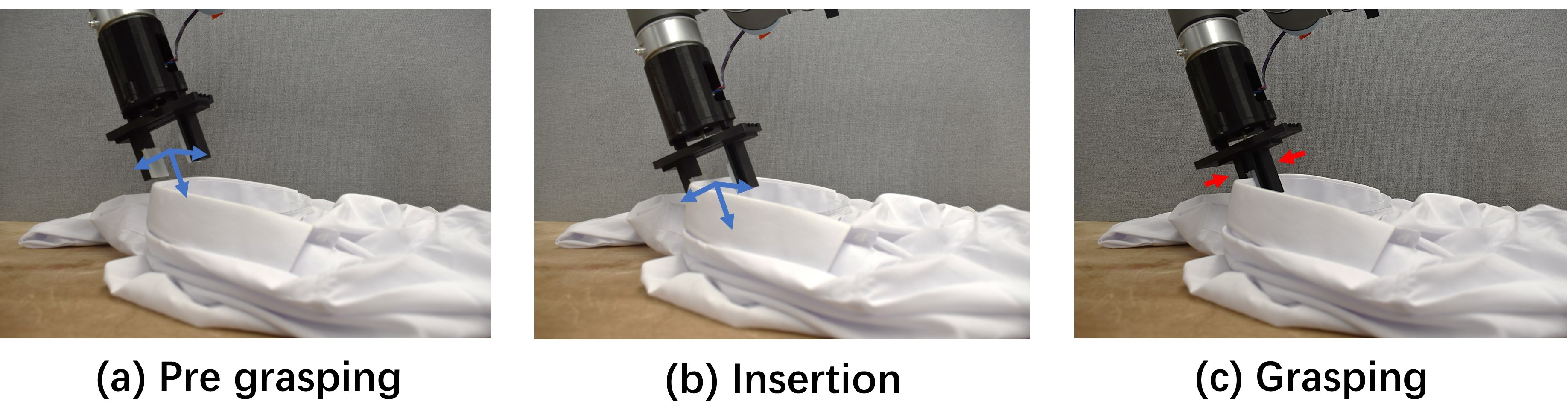}
    \caption{Insertion Trajectory: (a) The robot will approach to $(c_{pre}, o_{goal})$; (b) Insertion will execute robot to $(c_{goal}, o_{goal})$; (c) Grasping.}
    \label{fig:insertion}
    \vspace{-10pt}
\end{figure}

As shown in Fig. \ref{fig:insertion}, this study uses $(\bm{c},\bm{o})$ to specify the goal coordinate and orientation of the 6d grasping pose. We perform the calibration according to the known camera extrinsic parameters to transfer the pose of grasping region $(c_{goal}, o_{goal})$ to the world frame.
Based on the selection of the goal grasping pose, we define a pre-grasp pose $(c_{pre}, o_{goal})$, which is 50 mm away from the target position along the robot approaching direction ($Z$-axis). The robot arm will first reach the pre-grasping pose, then perform the grasp insertion, in order to make sure the collar of the garment is inserted between each gripper finger. Upon reaching the grasping pose, the robot performs the grasp. The motion planning of the robot is achieved by Moveit! \cite{chitta2016moveit}. 

\subsection{Garment-hanging}
Garment-hanging is challenging due to the significant variation in appearance and topology of garments. To approach this problem, we treat the grasping point as the keypoint that represents the whole configuration of cloth, in a similar method to \cite{manuelli2019kpam}. By doing so, we assume that if the collar is grasped successfully, then the garment will be in a suitable state to able to be hung on a peg without the need for further regrasping. As shown in Robot Execution of Fig. \ref{fig:data_acquisition}, after the robot grasps the cloth, the robot will execute a pre-defined hanging trajectory. After reaching the intermediate position (blue dot) and the center of the hook (green dot), the robot releases the garment and finishes the hanging task.

\begin{table}[t]
\centering

\caption{Grasping Success Rate with Robot and A Parallel Gripper.}
\vspace{-2mm}
\setlength{\tabcolsep}{2mm}{
\begin{tabular}{c|c c c c c}
\hline
&\multicolumn{5}{c}{Initial Configuration} \\
\hline
Method  &\textbf{[fd1]} & \textbf{[cr1]} &\textbf{[cr2]} & \textbf{[cr3]}& \textbf{[cr4]}  \\ [0.2ex]
\hline
\makecell[c]{*B1: Predicted-Point}       &48\% &28\%&22\% &20\% &16\%  \\
\hline
\makecell[c]{*B2: Predicted-Point-o}     &80\% &44\%&34\% &26\% &20\% \\
\hline
\makecell[c]{B3: Only-Centering}         &58\% &46\%&34\% &30\% &24\% \\
\hline
\makecell[c]{B4: Only-Surface-Variation} &80\%&66\%&56\%&38\% &28\% \\
\hline
\makecell[c]{Our Method}    &\textbf{92\%}&\textbf{80\%}&\textbf{62\%}&\textbf{50\%} &\textbf{36\%} \\
\hline
\end{tabular}
}
\par
\begin{flushleft}
\textbf{[fd1]}, \textbf{[cr1]}-\textbf{[cr4]} correspond to experiment scenarios where folded garment (1 piece), crumpled garments (1-4 pieces) are randomly placed on the tabletop, respectively. TS and S1 are used for grasping evaluation. The letter B indicates a baseline method. *Baseline 1 and *Baseline 2 are based on\cite{ramisa2012using,zhang2020learning}.
\end{flushleft}
\label{table:grasping}
\vspace{-10pt}
\end{table}

\section{Experiment}
Several experiments are conducted here to test our proposed cloth manipulation system. We evaluate the proposed cloth manipulation system in three aspects: 1) the accuracy and robustness of our proposed learned perception model in finding the collar of garments; 2) the performance of our proposed grasping strategy; 3) the overall evaluation of the robotic hanging system.
\subsubsection{Perception Evaluation}
In the experiment to evaluate the perception system, we use the method described in Section III-A to prepare the training set, validation set and testing set. The training set is comprised of 8212 depth images, the validation set of 2023 depth images with the same garment, and the testing set contains 5328 depth images from unseen garments. Our experiment uses template shirts (TS) with two long sleeves to collect training and validation sets. Short sleeve shirts (S1, S2), polo shirts (P1, P2), long sleeved shirts (L1, L2), coats (C1, C2) and denim jackets (D1, D2) are used here for the testing set. We use Intersection over Union (IoU), Recall and Precision to measure the performance of the system.
From Table \ref{tab:perception}, it can be observed that our proposed perception system can generalize to other collard garments. Specifically, our proposed system has a high accuracy in seen garments and most unseen garments. Although some accuracy gap between seen and unseen objects exists, the overall performance of unseen garments is still satisfactory for downstream manipulation tasks.
More positive and negative examples of garment segmentation are presented in the \href{https://sites.google.com/view/garment-hanging}{multimedia materials supplied with this work.}

\begin{table*}[t!]
    \centering
     \vspace{4mm}
    \caption{Performance of Robotic Hanging}
    \vspace{-2mm}
   \setlength{\tabcolsep}{3.5mm}{ 
   \begin{tabular}{c c c c c c c c c c c c c c c}
        \hline
        Initial Config. & TS & S1 & S2 & L1 & L2 & P1 & P2 & C1 & C2 & D1 & D2 & Overall\\ [0.1ex]
        \hline
        \textbf{[fd]}      &9/10 &8/10 &8/10 &7/10 &6/10 &2/10 &6/10 &8/10 & 7/10& 6/10& 6/10 & 73/110\\
        \textbf{[cr]}     &6/10 &7/10 & 6/10 &6/10 &6/10 &2/10 &5/10 &7/10 &5/10& 6/10& 5/10 & 61/110\\
            \hline
    \end{tabular}}
       \par
         \begin{flushleft}
To demonstrate the robustness of our proposed system in real-world scenarios, 11 different garments used in perception experiments are applied for this task. \textbf{[fd]} and \textbf{[cr]} represent folded and crumpled configuration of garments respectively.
\end{flushleft}
     \vspace{-18pt}
    \label{table:hanging}
\end{table*}

\subsubsection{Grasping Strategy Evaluation}
The real-world garment grasping is implemented on a UR5 robot with a customized parallel gripper. All gripper parts are 3d-printed, and the gripper is driven by a NEMA-17 stepper motor. A soft silicone pad is attached to each finger to ensure high friction during grasping. An Intel RealSense D435i RGB-D camera is fixed above the scene during both data collection and experiment. All training and inference is performed on a machine running Ubuntu 18.04 with an NVIDIA GTX 3070 GPU. 

A range of initial configurations of garments is essential to test the robustness of the grasping strategy. In grasping evaluation experiments, we follow the benchmark proposed by \cite{garcia2020benchmarking}. From this benchmark, two initial configurations: folded \textbf{[fd]} and crumpled \textbf{[cr]} are adopted. 
For the \textbf{[fd]} configuration, the garment is initialized with a relatively structured configuration. The target collar region of the garment is completely visible with minimal self-occlusion and deformation.
For \textbf{[cr]}, the garment is dropped from height of 0.4 meters above the tabletop at the start of each trial, resulting in a crumpled garment configuration. Generally, the target collar region of the garment is partially visible with self-occlusion and deformation. With more garments added for the experiment (\textbf{[cr1]} to \textbf{[cr4]}), the degree of self-occlusion and deformation increases.
For each scenario, 50 grasping trials are performed to obtain the grasping success rate. Success cases are only counted when the collar regions are grasped and lifted 0.4 meters above the tabletop.
In summary, our method uses a neural network to segment the collar region of garments, determines the center of the segmented collar pixels with a skeleton-based method, selects the pose aligning with the maximum surface variation to grasp. Several baselines are also implemented to evaluate the effectiveness of our proposed method and test the contribution of each aspect of the method:
\begin{itemize}
    \item
    \textbf{Predicted-Point} uses TSs with a tagged collar center for data acquisition to train a model that directly predicts the grasping point, similar to \cite{ramisa2012using, zhang2020learning}. We use the same architecture described in Section III-A for the model training. A \textbf{fixed} top-down grasping orientation is executed for the grasping.
    \item
    \textbf{Predicted-Point-o} uses the same point prediction model, but the grasping orientation estimation is determined by the method described in Section III-B-(3).
    \item
    \textbf{Only-Centering} uses the segmented collar of the shirt. The grasping point is randomly chosen within the center region. The grasping orientation is determined by the method described in Section III-B-(3).
    \item
    \textbf{Only-Surface-Variation} uses the segmented collar of the shirt. The point with the maximum surface variation of the whole segmentation output is selected as the grasping point. The grasping orientation is determined by using the method described in Section III-B-(3).
\end{itemize}

\begin{figure}[t!]
    \centering
    \vspace{4mm}
    \includegraphics[width=0.98\columnwidth]{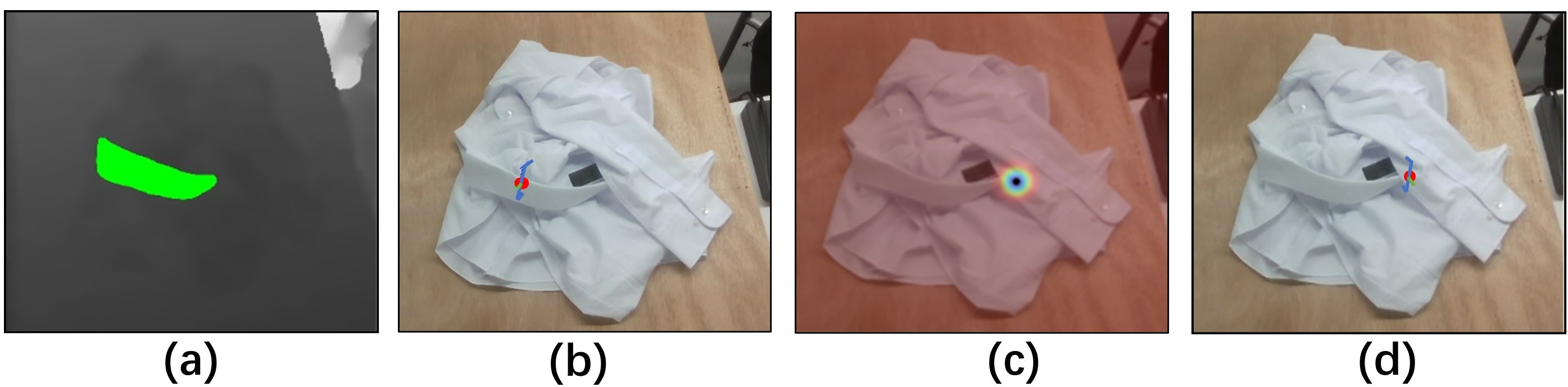}
    \caption{Segmentation and predicted point for collar grasping methods. The result of segmentation (a) and grasping estimation (b) of our method. Point detection (c) and grasping estimation (d) of Predicted-Point-o.
    }.
    \label{fig:baselines}
    \vspace{-10pt}
\end{figure}

As shown in Table \ref{table:grasping}, segmentation-based methods outperform methods that rely on image-to-grasping point prediction in unstructured (\textbf{[cr1]} - \textbf{[cr4]}) scenarios. This is largely because the trained target point in point prediction methods may not always be visible. This is illustrated further in Fig. \ref{fig:baselines}, which shows a failure case of point methods, where the target point is occluded. Full grasping pose estimation (Predicted-Point vs Predicted-Point-o) is shown to be critical for improved grasping performance, and is necessary when the initial configuration is not restricted to a single orientation. Identifying the collar fold is also important for successful grasping, with methods that identify maximum surface variation (the proposed method and the Only-Surface-Variation baseline) outperforming those that do not in all cases. Finally, combining centering and surface variation allows our method to ensure a good contact between gripper and garment, avoiding the extreme edges of the collar which lead to failure in the Only-Surface-Variation baseline.
The success rate of these methods decreases as more garments are added in the grasping scene. This strongly reduces segmentation performance because of more occlusion and an increased garment configuration complexity. However, even in the most complex scenario (\textbf{[cr4]}), the proposed method still outperforms other baselines.
 
\subsubsection{Manipulation System Evaluation}

We evaluate our proposed system based on a garment-hanging task.
As shown in Table \ref{table:hanging}, with our proposed vision-based perception and grasping algorithm, our system has 73 successful cases out of 110 trials for the \textbf{[fd]} configuration, while the \textbf{[cr]} realizes 61 cases - a slight decrease. 
Furthermore, we observed that failures of garment-hanging are mainly caused by: (1) failed initial grasping; (2) garment dropping because of heavy weight; (3) garment entangled during the hanging. These are shown in the \href{https://sites.google.com/view/garment-hanging}{multimedia material.} Of these, (1) represents a failure case regarding the proposed algorithm, while (2) can be addressed with a stronger gripper, and (3) may require a more sophisticated hanging policy as opposed to the offline policy used here.

\section{Conclusions and Future Work}

In this paper, we propose a novel cloth manipulation system that utilises the structural elements of garments for both perception and grasping. The training of this perception system is based on data acquisition of a 10-minute video of human manipulation with automatic annotation. Experimental results indicate the proposed perception system can generalize to other types of similarly structured clothes. With the proposed pose estimation strategy, our system can achieve 92\% and 80\% grasping success rates for a single piece of folded and crumpled garment, respectively, which are significantly higher than other baselines. We use a garment-hanging task as a case study to evaluate the performance of our overall system. Results indicate that the challenging garment-hanging task can be implemented with an offline hanging policy by only considering the structural regions of garments used in grasping.
Future work will investigate how to explore and search scene in situations where garment collar is completely non-visible. More work will be done by using our proposed system to perceive and grasp other vital regions (e.g. hem or sleeves) of garments for implementing different garment manipulation tasks.

\addtolength{\textheight}{-1cm}   


\bibliographystyle{IEEEtran}
\bibliography{main}   

\end{document}